# Quotient Based Multiresolution Image Fusion of Thermal and Visual Images Using Daubechies Wavelet Transform for Human Face Recognition

Mrinal Kanti Bhowmik[1], Debotosh Bhattacharjee[2], Mita Nasipuri[2], Dipak Kumar Basu[2*] and Mahantapas Kundu[2]

[1]Department of Computer Science and Engineering, Tripura University (A Central University)
Suryamaninagar, Tripura 799130, India
*mkb_cse@yahoo.co.in*

[2] Department of Computer Science and Engineering, Jadavpur University
Kolkata, West Bengal 700032, India
*AICTE Emeritus Fellow
*debotosh@indiatimes.com, mitanasipuri@gmail.com, dipakkbasu@gmail.com, mkundu@cse.jdvu.ac.in*

**Abstract**
This paper investigates the multiresolution level-1 and level-2 Quotient based Fusion of thermal and visual images. In the proposed system, the method-1 namely "Decompose then Quotient Fuse Level-1" and the method-2 namely "Decompose-Reconstruct then Quotient Fuse Level-2" both work on wavelet transformations of the visual and thermal face images. The wavelet transform is well-suited to manage different image resolution and allows the image decomposition in different kinds of coefficients, while preserving the image information without any loss. This approach is based on a definition of an illumination invariant signature image which enables an analytic generation of the image space with varying illumination. The quotient fused images are passed through Principal Component Analysis (PCA) for dimension reduction and then those images are classified using a multi-layer perceptron (MLP). The performances of both the methods have been evaluated using OTCBVS and IRIS databases. All the different classes have been tested separately, among them the maximum recognition result is 100%.

## 1. Introduction

Face recognition is a vital problem in computer vision. Though face recognition systems show considerable improvement in successive competitions [1] [2], still it is considered unsolved [3]. Face recognition needs high degree of accuracy as its most applications are in public security, law enforcement and commerce, such as mug-shot database matching, identity authentication for credit card or driver license, access control, information security and intelligent surveillance [4]. There are a lot of factors (like, illumination variation, pose variation, facial expression changes etc) which affect the face recognition performance. Among all these, illumination problem has received much attention in recent years. Quotient imaging technique is one of the solution to this problem. This method is simple and significant. Quotient Image is an image ratio between a test image and a linear combination of three images illuminated by non- coplanar lights, depends only on the albedo information, and therefore is illumination free [5]. The quotient image can be considered as fused quotient image as both the visual and its corresponding thermal images have been used to generate it.

The processing steps of the two methods used to generate quotient images are shown in the Fig. 1(a) and Fig. 1(b). In both the two methods for image decomposition purpose discrete 2-D wavelet transform has been used for both the visual and thermal images. But in method-2 for reconstruction purpose the inverse discrete 2-D wavelet transform has been used. In method-1, decomposition has been done at level-1. To generate the fused quotient image, all the coefficients of the decomposed image have been used. In case of method-2, discrete 2-D wavelet transform has been used at level-2 to decompose the visual and thermal images. But to regenerate the images only approximation coefficients have been used. With the new reconstructed visual and thermal images the quotient images have been generated.

In terms of designing a facial recognition system with high accuracy level, the main crucial point is the choice of feature extractor. In this connection, principal component analysis (PCA) has been used for dimension reduction purpose. Principal component analysis (PCA) is based on the second-order statistics of the input image, which tries to attain an optimal representation that minimizes the reconstruction error in a least-squares sense. Eigenvectors of the covariance matrix of the face images constitute the eigenfaces. The dimensionality of the face feature space is reduced by selecting only the eigenvectors possessing significantly large eigenvalues [25]. The eigenfaces which are the set of eigenvectors is then used to describe face images [6]. These eigenfaces are then classified using Multi-layer perceptron (MLP).

Different existing illumination invariant methods have been discussed. Researchers have proposed different solutions to illumination problem, which include invariant feature based method [7], 3D linear illumination subspace method [8], linear object class [9], illumination and pose manifold [10], illumination cones [13], Harmonic subspace [17], lambertian reflectance and linear subspace [14] and individual PCA combining the synthesized images [15] [16]. Theoretically the Illumination Cone method illustrated that face images due to varying lighting directions form an illumination cone [1]. In this algorithm, both self-shadow and cast-shadow were considered and its experimental results outperformed most existing methods. Ramamoorthi [17-19] and Basri [20] [21] independently developed the spherical harmonic representation. This original representation explained why the images of an object under different lighting conditions can be described by low dimensional subspace in some previous empirical experiments [23] [24]. Among all these algorithms the quotient image method is simple and practically useful also. Quotient image by Shashua and Riklin-Raviv is mainly designed for dealing with illumination changes in face recognition [26] [27] [28]. Jacobs et. al. [30] introduced another concept of quotient image, which is the ratio of two images. Jacobs's method considers the Lambertian model without shadow and assumes the surface of the object is smooth [5]. Retinex, which is a combination of the words retina and cortex, is an algorithm to model the human visual system [22]. Though its original purpose is for color constancy, but performs well as a contrast-enhancement algorithm too. A famous algorithm called Frankle-McCann variation has been introduced by McCann and Frankle [33] and this algorithm works by operating on the image pixels in the log domain. This involves four basic steps: ratio, product, reset and average [29]. In this approach the main contribution is the generation of fused quotient images (I/J) from both the coefficients of visual (I) and thermal (J) images. In this paper, first time the authors have contributed the quotient based multiresolution image fusion of thermal and visual images using Daubechies Wavelet Transform for human face recognition.

The paper is organized as follows: System Overview has been given in Section-2 which includes the description of visual and thermal face images, multiresolution analysis, the image decomposition and reconstruction process, quotient imaging method, Principal Component Analysis (PCA) and Artificial Neural Network. Section-3 shows and analyses the experimental results using OTCBVS and IRIS databases. The comparison between different quotient imaging methods are shown in Section 4 and finally the conclusion is made in section 5.

## 2. System Overview

Here, a technique for human face recognition using quotient images has been present. The block diagram of the system is given in Fig. 1(a) for method-1 and in Fig. 1(b) for method-2. All the processing steps to generate quotient images used in two methods are shown in the corresponding block diagrams. In case of method-1 in first step single level decomposition has been done for both the visual and thermal images. Quotient images are generated from all the coefficients (approximation and details coefficients) of decomposed visual and thermal images. Method-2 is slightly different from method-1. In the first step, decomposition of both the thermal and visual images up to level-2 has been done using Daubechies wavelet. The reason behind using level-2 decomposition is that the higher the decomposition levels are the more advantageous because the number of higher and lower frequency subbands will increase. In this connection the size of the image will decrease and thus the processing speed will increase [38]. Then the images are reconstructed from the corresponding approximation coefficients in case of both visual and thermal images. Then Quotient images are generated with these coefficients.

These transformed images are separated into two groups namely training set and testing set. The PCA is computed using training images. All the training and testing images are projected into the created eigenspace and named as quotient fused eigenfaces. Once these conversions are done the next task is to use Multilayer Perceptron (MLP) to classify them. A multilayer feed forward network is used for classification.

2.1 Multiresolution Analysis

In computer vision, it is difficult to analyze the information content of an image directly from the gray-level intensity of the image pixels. Indeed, this value depends upon the lighting conditions. Generally, the structures we want to recognize have very different sizes. Hence, it is not possible to define a priori an optimal resolution for analyzing images.

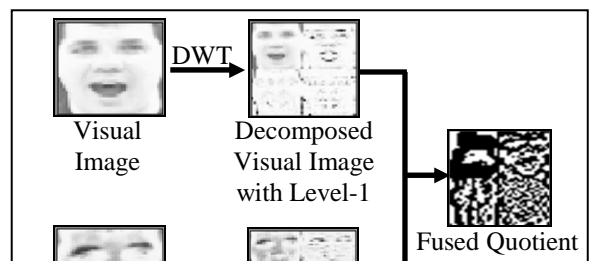

Fig. 1 Block Diagram of the System Presented Here for (a) method- 1 and (b) method- 2.

Wavelet decomposition is the most widely used multiresolution technique in image processing [34]. In this work, 2D discrete wavelet transform has been used to extract multiple subband face images. These subband images contain approximation coefficients matrix and details coefficients matrices like horizontal, vertical and diagonal coefficients of faces at various scales. One-level wavelet decomposition of a face image is shown in Fig. 2:

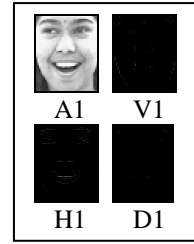

Fig. 2 Sample one-level wavelet decomposed image.

In Fig. 2, for both the methods A1 is the approximation coefficient, V1, H1 and D1 are the vertical, horizontal and diagonal details coefficients respectively. Single level decomposition is used in both visual and thermal images for method-1. An 80 × 100 pixels image is taken as input and after decomposition four 40 × 50 pixels resolution sub band images-A1, H1, V1 and D1 are obtained. In case of method-2, both the input images (visual and thermal) are of size 80×100. After single level decomposition, all the generated coefficients (A1, H1, V1 and D1) are of size 40×50. Again decomposition at level-2 has been applied only in approximation coefficient (A1) and after that, using only approximation coefficient, getting after level-2 decomposition, the quotient images of size 40×50 has been generated.

2.2 Daubechies Wavelet Transform

In this work Daubechies (db1) wavelet has been used to decompose the images as well as to reconstruct the images also. Daubechies (db1) wavelet is the same as Haar wavelet. So the discussion of the Haar wavelet is essential to understand the concept of Daubechies (db1) wavelet. In mathematics, the Haar wavelet is a certain sequence of functions. This sequence was proposed in 1909 by Alfréd Haar. Haar used these functions to give an example of a countable orthonormal system for the space of square-integrable functions on the real line [18]. The Haar wavelet's mother wavelet function $\psi(t)$ can be described as

$$\psi(t) = \begin{cases} 1 & 0 \leq t < 1/2, \\ -1 & 1/2 \leq t < 1, \\ 0 & otherwise. \end{cases} \quad (1)$$

and its scaling function φ (t) can be described as

$$\phi(t) = \begin{cases} 1 & 0 \leq t < 1, \\ 0 & otherwise. \end{cases} \quad (2)$$

Daubechies wavelet transform has been used in this work. Daubechies wavelets are a family of orthogonal wavelets defining a discrete wavelet transform and characterized by a maximal number of vanishing moments for some given support. This kind of 2-dimensional Discrete Wavelet Transform (DWT) aims to decompose the image into approximation coefficients (cA) and detailed coefficients cH, cV and cD (horizontal, vertical and diagonal) obtained by wavelet decomposition of the input image (X). The 2-dimensional Discrete Wavelet Transform (DWT) functions used in Matlab-7 are shown in equation (3) and equation (4).

[cA, cH, cV, cD] = dwt2 (X, 'wname')             (3)
[cA, cH, cV, cD] = dwt2 (X, Lo_D, Hi_D)          (4)

In Eq. (3), 'wname' is the name of the wavelet used for decomposition. In this work db1 has been used in case of wname. Eq. (4) Lo_D (decomposition low-pass filter) and Hi_D (decomposition high-pass filter) wavelet decomposition filters.

This kind of two-dimensional DWT leads to a decomposition of approximation coefficients at level j in four components: the approximation at level j+1, and the details in three orientations (horizontal, vertical, and diagonal). The Fig. 3 describes the algorithmic basic decomposition steps for image where, a block with a down-arrow indicates down-sampling of columns and rows and cA, cH, cV and cD are the coefficient vectors [37] [38] [39].

Consequently the reconstruction process is performed using inverse of DWT (IDWT). We have reconstructed the images based on the approximation co-efficient matrix cA. Finally the reconstructed image is used as the input to PCA for dimension reduction. The 2-dimensional Inverse Discrete Wavelet Transform (IDWT) functions used in Matlab-7 are shown in equation (5) and equation (6)

X = idwt2 (cA, cH, cV, cD, 'wname')             (5)
X = idwt2 (cA, cH, cV, cD, Lo_R, Hi_R)          (6)

Inverse discrete wavelet transform (IDWT) uses the wavelet 'wname' to compute the single-level reconstruction of an Image X, based on approximation matrix (cA) and detailed matrices cH, cV and cD (horizontal, vertical and diagonal respectively). In the Fig. 4 we have shown the algorithmic basic reconstruction steps for an image respectively.

The above concepts of discrete wavelet transform (DWT) function and inverse discrete wavelet transform (IDWT) function have been used from wavelet toolbox of Matlab-7.

Fig. 3 Steps for decomposition of an image for method-1 and method-2.

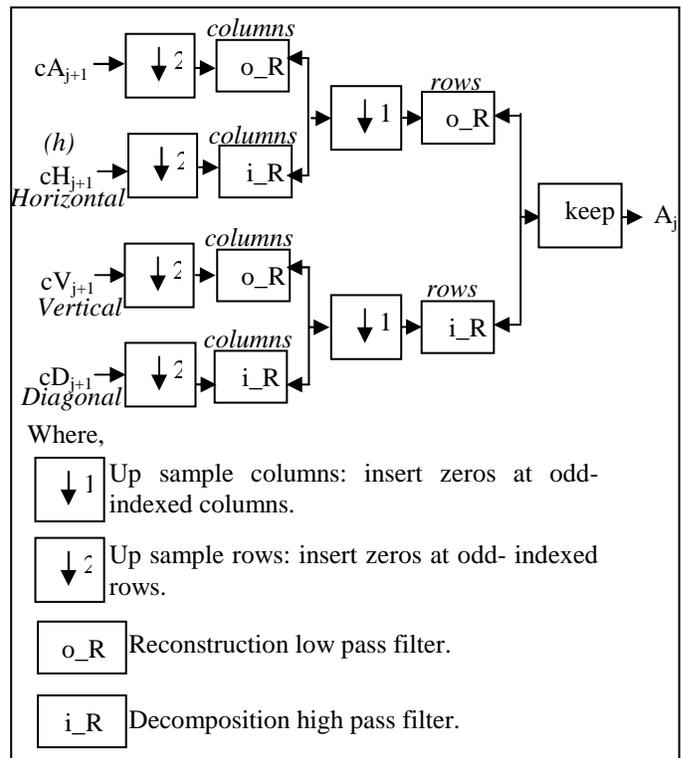

Fig. 4 Steps for reconstruction of an image for method-2.

### 2.3 The Quotient Fused Image of Thermal and Visual Image

If two objects are 'a' and 'b', we define the quotient image Q by the ratio of their albedo functions $\rho_a$ and $\rho_b$. Clearly, Q is illumination invariant. In the absence of any direct access to the albedo functions, it has been shown that Q can nevertheless be recovered, analytically, given a bootstrap set of images. Once Q is recovered, the entire image space (under varying lighting conditions) of object 'a' can be generated by Q and three images of object 'b'. The details are given below. Let, start with the case N = 1, i.e., there is a single object (2 images) in the bootstrap set. Let the albedo function of that object 'a' be denoted by $\rho_a$, and let the two images be denoted by $a_1$, $a_2$. Therefore, $a_j = \rho_a n^T s_j$, where, j = 1,2. Let y be another object of the class with albedo $\rho_y$ and let $y_s$ be an image of y illuminated by some lighting condition s, i.e., $y_s = \rho_y n^T s$. The quotient image $Q_y$ of object y against object a is defined by

$$Q_y(u,v) = \frac{\rho_y(u,v)}{\rho_a(u,v)} \quad (7)$$

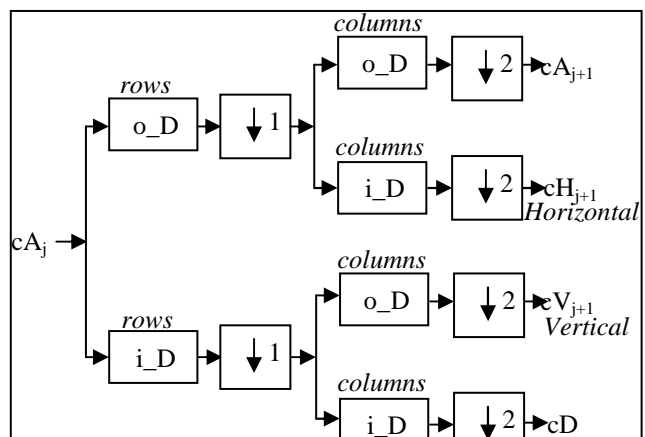

Where u, v range over the image.
Thus, the image $Q_y$ depends only on the relative surface texture information, and thus is independent of illumination [40].The quotient image used in this experiment is the quotient of the visual and thermal image of an object. Let, the visual image be I and the thermal image be J, then we can consider the quotient image C as f (I) / f (J). The above mentioned quotient imaging technique is same for both the methods (method-1 and method-2). But f (I) and f (J) are different for method-1 and method-2. For method-1, first both the visual and thermal images have been decomposed at level-1. The quotient images have been generated using all the four coefficients of the decomposed images (approximation coefficient and details coefficients in three orientations (horizontal, vertical, diagonal)). In case of method-2, first the visual and thermal images are decomposed at level-2 and after that both the images are reconstructed using the approximate coefficients. These images are decomposed and reconstructed using wavelet decomposition and reconstruction function. Finally the quotient image is generated from both the reconstructed visual and thermal functions.

2.4 Image Fusion Rules

At the time of generating fused images of coefficients (approximate and details), the absolute maximum of thermal and visual images was selected The fusion method used to generate fused image of wavelet toolbox is present in the eq. (8).

$$D = abs(T) \geq abs(V)$$
$$C = T(D) + V(\sim D) \quad (8)$$

Where, D is the absolute maximum matrix of T (thermal image) and V (visual image), a*bs (T)* and *abs (V)* are the absolute matrices of T and V and C is the generated fused image of T and V.

Now at the time to find the absolute value of corresponding thermal and visual coefficients it will calculate the intensity of the each pixel. In case of equation 8, it will check that, $(T_{xy}) \geq (B_{xy})$, then for the fused image of coefficient it will do the following operation given in eq. (9):

$$F_{xy} = T_{xy} + V_{xy} \quad (9)$$

Where $F_{xy}$ is the pixel value of the fused image (C) and the method shown in eq. (9) is the process to calculate the pixel value for fused image of T and V.

2.5 Comparison between QI and SQI

In this section, a comparison analysis of quotient image (QI) and self-quotient image (SQI) has been presented. According to the concept proposed by Haitao Wang, Stan Z Li and Yangsheng Wang [25] SQI is defined as the ratio of the input image and its smooth versions. The self-quotient image Q of image I can be expressed as

$$Q = \frac{I}{\hat{I}} = \frac{I}{F * I} \quad (10)$$

where $\hat{I}$ is the smoothed version of I. F is the smoothing kernel. We can consider the SQI as one kind of QI which is derived from the same image I itself.

The definition of the quotient image provides an invariant representation of face images under different lighting conditions.

2.6 Principal Component Analysis

The Principal Component Analysis (PCA) [41], [42], [43] uses the entire image to generate a set of features and does not require the location of individual feature points within the image. We have implemented the PCA transform as a reduced feature extractor in our face recognition system. Here, each of the quotient fused face images are projected into the eigenspace created by the eigenvectors of the covariance matrix of all the training images represented as column vector. Here, we have taken the number of eigenvectors in the eigenspace as 40, because eigenvalues for other eigenvectors are negligible in comparison to the largest eigenvalues [44] [45] [46].

2.7 ANN using Back propagation with Momentum

Neural networks, with their remarkable ability to derive meaning from complicated or imprecise data, can be used to extract patterns and detect trends that are too complex to be noticed by computer techniques. A trained neural network can be thought of as an "expert" in the category of information it has been given to analyze. The Back propagation learning algorithm is one of the most historical developments in Neural Networks. It has reawakened the scientific and engineering community to the modeling and processing of many quantitative phenomena using neural networks. This learning algorithm is applied to multilayer feed forward networks consisting of processing elements with continuous differentiable activation functions. Such networks associated with the back propagation learning algorithm are also called back propagation networks [47] [12] [25] [35].

## 3. Experiments and Discussion

Experiments are performed to evaluate Quotient Image (QI) for face recognition, using IRIS database. This work has been simulated using MATLAB 7 in a machine of the configuration 2.13GHz Intel Xeon Quad Core Processor and 16384.00MB of Physical Memory. As the both reconstructed visual and thermal images has been used to generate quotient fused images, so first of all original images are cropped manually.

### 3.1 OTCBVS Database

The experiments were performed on the face database which is Object Tracking and Classification Beyond Visible spectrum (OTCBVS) benchmark database containing a set of thermal and visual face images. This is a publicly available benchmark dataset for testing and evaluating novel and state-of-the-art computer vision algorithms. The benchmark contains videos and images recorded in and beyond the visible spectrum; it contains different sets of data like: OSU Thermal Pedestrian Database, IRIS Thermal/Visible Face Database, OSU Color-Thermal Database, Terravic Facial IR Database, Terravic Weapon IR Database and CBSR NIR Face Dataset. Among all of these different datasets, mainly, visual images from IRIS Thermal/Visible Face Dataset have been used. In this dataset, there are 2000 images of visual and 2000 thermal images of 16 different persons. In case of method-1, the total 770 visual images and their corresponding thermal images of 14 different classes were choosen. Randomly chosen 356 images from 14 different classes are used for training purpose and 242 images from 8 different classes are used for testing purpose. In case of method-2, total number of training images and total number of testing images are same, which is 110 and the number of classes is 10. For some subject, the images were taken at different times which contain quite a high degree of variability in lighting, facial expression *(open /* closed eyes, smiling /non smiling etc.), pose (Up right, frontal position etc.) and facial details (Glasses/ no Glasses). All the images were taken against a dark homogeneous background with the subjects in upright, fontal position, with tolerance for some tilting and rotation of up to 20 degree. The variation in scale is up to about 10% for all the images in the database [11]. Some sample visual and thermal images and their corresponding quotient fused images are shown in Fig. 5:

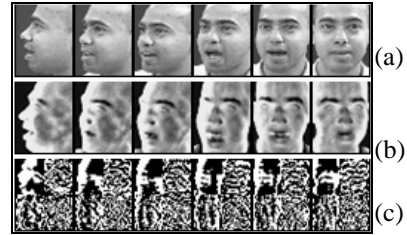

Fig. 5 Some sample (a) Visual Images (b) Thermal Images and (c) Corresponding Quotient Fused Images of OTCBVS Database used for quotient imaging method 1.

### 3.2 IRIS Thermal/Visual Face Database

In this database, all the thermal and visible unregistered face images are taken under variable illuminations, expressions, and poses. The actual size of the images is 320 x 240 pixels (for both visual and thermal). 176-250 images per person, 11 images per rotation (poses for each expression and each illumination). Total 30 classes are present in that database and the size of the database is 1.83 GB [28]. In this experiments total 246 visual and 246 thermal images from this database of 8 different classes has been used for method-1.Total number of quotient images generated for method-1 is 246. Among this 246 images, randomly selected, 82 images are used as training data and 164 images are used as testing data. 220 visual and 220 thermal images from this database of 10 different classes have been used to evaluate the performance of method-2. Number of quotient images are 220. So, in case of method-2, the total number of visual and thermal images used in this work is 440 and the total number of quotient images used in the experiment is 220. Randomly selected 110 quotient images are used as training data and 110 quotient images are used as testing data of different 10 classes. Training has been conducted with all the 10 classes, but different 10 classes have been tested separately to evaluate the performance of all the classes separately. Some sample visual and thermal images and their corresponding quotient images for method-1 and method-2 are shown in Fig. 6 and Fig. 7 respectively.

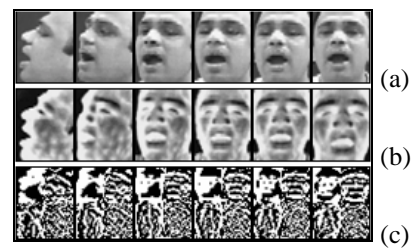

Fig. 6 Some sample (a) Visual Images (b) Thermal Images and (c) Corresponding Quotient Fused Images of IRIS Database used for quotient imaging method-1.

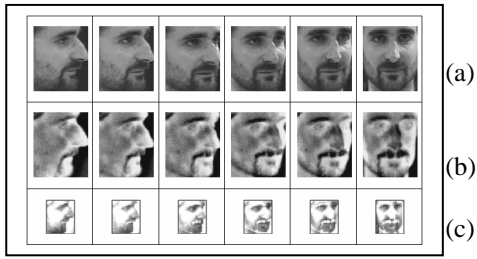

Fig. 7 Some sample (a) Visual Images (b) Thermal Images and (c) Corresponding Quotient Images of IRIS Database used in Quotient Imaging Method-2.

3.3 Training and Testing of the Experiments:

In case of method-1, quotient fused images have been used. Experiments have been conducted on OTCBVS and IRIS databases. At first, the original visual and thermal images of both the databases have been cropped and resized into 40 x 50 pixels. After resizing all the images, quotient-fused images have been generated from their corresponding thermal-quotient and visual-quotient images. All the training data of both the experiments using IRIS and OTCBVS databases are shown in Fig. 8. From Fig. 8 it can be find out that 8 different training image classes of IRIS database and 14 different training classes of OTCBVS database have been used. Total number of images of IRIS and OTCBVS databases used as training images is 82 and 356 respectively. Number of images per class is not same for all the classes.

For method-2, experiments have been used only on IRIS database. First, the original visual and thermal images have been cropped and resized into 80 x 100 pixels. After resizing all the images, the quotient images of size $40 \times 50$ pixels have been generated from their corresponding thermal coefficient and visual coefficient images.

Some sample training data of the experiment are shown in Fig. 9. The number of images, taken per class is 10 for both the training set and testing set. Training has been done with the whole training dataset. But the different classes have been tested separately. The average recognition rate for all the classes is 91%. But for 3 classes (class-1, class-3 and class-6) it reaches the maximum recognition level i.e. 100%. Total number of training classes is 10, and the number of images per class is 11. So the total number of training data is 110.

Fig. 8 Training data (Quotient Image) of the Experiments using IRIS and OTCBVS Databases.

All the experimental results using OTCBVS and IRIS databases for method-1 are shown in table 1 and for method-2 are shown in table 2.

In case of method-1 using IRIS database, total number of testing classes is 8. The highest recognition rate is 100% which we got for Class-1, Class-2 and Class-3 of IRIS database testing classes. In case of OTCBVS the total number of testing classes is 8 and the highest recognition rate is 100% for class-8.

In case of method-2, number of testing images per class is same for all the classes i.e. 11. But the number of recognized images is not same for all the classes, so we got different recognition rates for different classes.

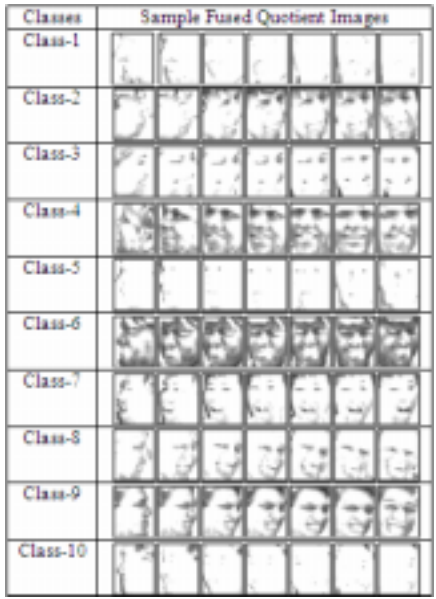

Fig. 9 Training data (Fused Quotient) of the Experiments using IRIS Databases.

Table 1: Experimental Results of Method-1 using OTCBVS and IRIS Database.

| Classes | OTCBVS | | IRIS | |
|---|---|---|---|---|
| | *Total no. of Testing/ Recognized Images* | *Recognition Rate* | *Total no. of Testing/ Recognized Images* | *Recognition Rate* |
| Class 1 | 33/32 | 97% | 22/22 | 100% |
| Class 2 | 33/30 | 91% | 22/22 | 100% |
| Class 3 | 22/20 | 91% | 20/20 | 100% |
| Class 4 | 33/30 | 91% | 20/18 | 90% |
| Class 5 | 22/20 | 91% | 20/18 | 90% |
| Class 6 | 33/31 | 94% | 20/17 | 85% |
| Class 7 | 33/31 | 94% | 20/18 | 90% |
| Class8 | 33/33 | 100% | 20/19 | 95% |

For class-1, Class-4 and Class-6, the recognition rate is maximum (100%) in case of IRIS database, so the false rejection rate for those classes is 0%.

The graphical representation of the testing result of method-1 and method-2 is shown in Fig. 10 and Fig. 11 respectively. The horizontal and vertical axes of these two graphs are for testing class's numbers and their corresponding recognition rates respectively.

In case of method-1, the system is trained by 352 images of IRIS database and 185 images of OTCBVS database. Total 110 images from each of the OTCBVS and IRIS databases have been used as training data for method-2.

Table 2: Experimental Results of Method-2 using OTCBVS and IRIS Database.

| Classes | OTCBVS | | IRIS | |
|---|---|---|---|---|
| | *Total no. of Testing/ Recognized Images* | *Recognition Rate* | *Total no. of Testing/ Recognized Images* | *Recognition Rate* |
| Class 1 | 11/9 | 82% | 11/11 | 100% |
| Class 2 | 11/11 | 100% | 11/10 | 91% |
| Class 3 | 11/9 | 82% | 11/10 | 91% |
| Class 4 | 11/8 | 73% | 11/11 | 100% |
| Class 5 | 11/10 | 91% | 11/10 | 91% |
| Class 6 | 11/8 | 73% | 11/11 | 100% |
| Class 7 | 11/11 | 100% | 11/10 | 91% |
| Class8 | 11/11 | 100% | 11/10 | 91% |
| Class 9 | 11/10 | 91% | 11/9 | 82% |
| Class 10 | 11/10 | 91% | 11/8 | 73% |

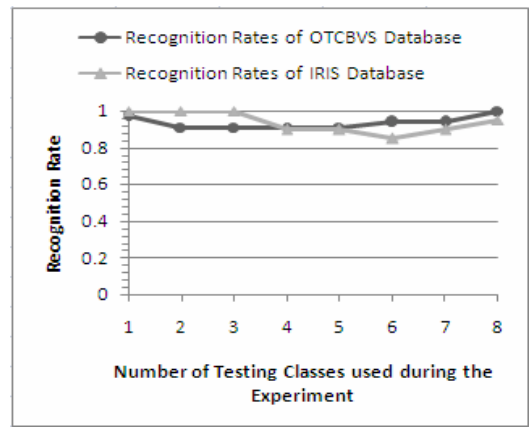

Fig. 10: Recognition Rate of Different Testing Classes of OTCBVS and IRIS Databases using Quotient Imaging Method- 1.

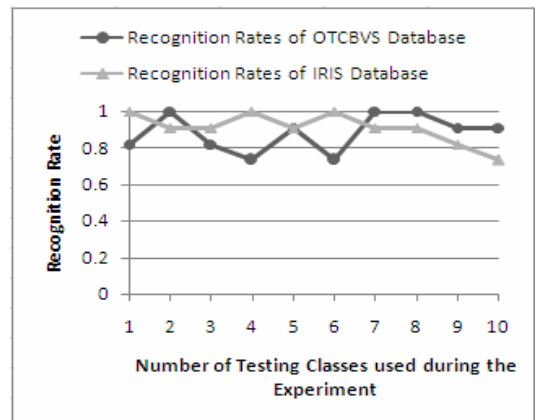

Fig. 11: Recognition Rate of Different Testing Classes of OTCBVS and IRIS Databases using Quotient Imaging Method-2.

## 4. Comparison between Different Quotient Imaging Methods

A comparative data of recognition rates of different quotient imaging method is shown in table 3.

Table 3: Comparison between Different Methods of Face Recognition.

| *Method* | *Recognition Rate* |
|---|---|
| Present Method (Method-1) + IRIS | 94% (average) 100% (maximum) |
| Present Method (Method-1) + OTCBVS | 94% (average) 100% (maximum) |
| Present Method (Method-2) + IRIS | 91% (average) 100% (maximum) |
| Present Method (Method-2) + OTCBVS | 88% (average) 100% (maximum) |
| CAQI-other [31] | 93% |
| CAQI-same [31] | 94% |
| SQI [31] | 70% |

## 5. Conclusions

An approach of face recognition using quotient image is presented here. The efficiency of our scheme has been demonstrated on Object Tracking and Classification Beyond Visible spectrum (OTCBVS) benchmark database and Imaging, Robotics and Intelligent Systems (IRIS) database, which contain images gathered with varying lighting, facial expression, pose, and facial details. The experimental results show that our method can significantly perform the face recognition task for the face images under different lightning conditions.

**Acknowledgments**

Mr. M. K. Bhowmik is thankful to the project entitled "Development of Techniques for Human Face Based Online Authentication System Phase-I" sponsored by Department of Information Technology under the Ministry of Communications and Information Technology, New Delhi-110003, Government of India Vide No. 12(14)/08-ESD, Dated 27/01/2009 at the Department of Computer Science & Engineering, Tripura University-799130, Tripura (West), India for providing the necessary infrastructural facilities for carrying out this work. The first author would also like to thank Prof. B.K. De, Dean of Science of Tripura University (A Central University), for their kind support to carry out this research work.